\def\year{2019}\relax
\documentclass[letterpaper]{article} 
\usepackage{aaai20}  
\usepackage{times}  
\usepackage{helvet} 
\usepackage{courier}  
\usepackage[hyphens]{url}  
\usepackage{graphicx} 
\usepackage{graphicx}  
\usepackage{booktabs}
\usepackage{arydshln}

\usepackage{amsthm} 
\usepackage{amsmath} 
\usepackage{amsfonts} 
\usepackage[mathscr]{eucal} 

\usepackage{subcaption}

\usepackage{algorithm}
\usepackage{algorithmic}

\def \inv {^{-1}}
\def \pHat {\hat{P_\pi}}
\def \pTilde {\tilde{P_\pi}}

\def \PsiHat {\hat{\Psi}_\pi}
\def \PsiTilde {\tilde{\Psi}_\pi}

\newtheorem{lemma*}{Lemma}
\newtheorem{theorem}{Theorem}
\newtheorem{theorem*}{Theorem}
\newtheorem{definition}{Definition}[section]

\title{Count-Based Exploration with the Successor Representation}
\author{Marlos C. Machado,\textsuperscript{\rm 1} \Large \textbf{Marc G. Bellemare,\textsuperscript{\rm 1}} \Large \textbf{Michael Bowling \textsuperscript{\rm 2}\textsuperscript{\rm 3}}\\ 
\textsuperscript{\rm 1} Google AI, Brain Team, \textsuperscript{\rm 2}University of Alberta,
\textsuperscript{\rm 3}DeepMind Alberta\\
\{marlosm, bellemare, bowlingm\}@google.com 
}
 
\begin{document}

\maketitle

\begin{abstract}
In this paper we introduce a simple approach for exploration in reinforcement learning (RL) that allows us to develop theoretically justified algorithms in the tabular case but that is also extendable to settings where function approximation is required. Our approach is based on the successor representation (SR), which was originally introduced as a representation defining state generalization by the similarity of successor states. Here we show that the norm of the SR, while it is being learned, can be used as a reward bonus to incentivize exploration. In order to better understand this transient behavior of the norm of the SR we introduce the substochastic successor representation (SSR) and we show that it implicitly counts the number of times each state (or feature) has been observed. We use this result to introduce an algorithm that performs as well as some theoretically sample-efficient approaches. Finally, we extend these ideas to a deep RL algorithm and show that it achieves state-of-the-art performance in Atari 2600 games when in a low sample-complexity regime.
\end{abstract}

\section{Introduction}
Reinforcement learning (RL) tackles sequential decision making problems by formulating them as tasks where an agent must learn how to act optimally through trial and error interactions with the environment. The goal in these problems is to maximize the (discounted) sum of the numerical reward signal observed at each time step. Because the actions taken by the agent influence not just the immediate reward but also the states and associated rewards in the future, sequential decision making problems require agents to deal with the trade-off between immediate and delayed rewards. Here we focus on the problem of exploration in RL, which aims to reduce the number of samples (i.e., interactions) an agent needs in order to learn to perform well in these tasks when the environment is initially unknown.

Surprisingly, the most common approach in the field is to select exploratory actions uniformly at random, with even high-profile success stories being obtained with this strategy (e.g., \citeauthor{Tesauro95} \citeyear{Tesauro95}; \citeauthor{Mnih15} \citeyear{Mnih15}). However, random exploration often fails in environments with sparse rewards, that is, environments where the agent observes a reward signal of value zero for the majority of states. In this paper we introduce an approach for exploration in RL based on the successor representation (SR; \citeauthor{Dayan93} \citeyear{Dayan93}). The SR is a representation that generalizes between states using the similarity between their successors, that is, the states that follow the current state given the agent's policy. The SR is defined for any problem, it can be learned with temporal-difference learning and, as we discuss below, it can be seen as implicitly estimating the transition dynamics of the environment.

The main contribution of this paper is to show that \emph{the norm of the SR can be used as an exploration bonus}. We perform an extensive empirical evaluation to demonstrate this and we introduce the substochastic successor representation~(SSR) to also understand, theoretically, the behavior of such a bonus. The SSR behaves similarly to the SR but it is more amenable to theoretical analyses. We show that the SSR implicitly counts state visitation, suggesting that the exploration bonus obtained from the SR, while it is being learned, might also be incorporating some notion of state visitation counts. We demonstrate this intuition empirically and we use this result to introduce algorithms that, in the tabular case, perform as well as traditional approaches with PAC-MDP guarantees. Finally, we extend the idea of using the norm of the SR as an exploration bonus to the function approximation case, designing a deep RL algorithm that achieves state-of-the-art performance in hard exploration Atari 2600 games when in a low sample-complexity regime. The proposed algorithm is also simpler than traditional baselines such as pseudo-count-based methods because it does not require domain-specific density models \cite{Bellemare16,Ostrovski17}.

\section{Preliminaries}
\label{sec:background}

We consider an agent interacting with its environment in a sequential manner. Starting from a state $S_0 \in \mathscr{S}$, at each step the agent takes an action $A_t \in \mathscr{A}$, to which the environment responds with a state $S_{t+1} \in \mathscr{S}$ according to a transition function $p(s'|s,a) = \mbox{Pr}(S_{t+1} = s'|S_t = s, A_t = a)$, and with a reward signal $R_{t+1} \in \mathbb{R}$, where $r(s,a)$ indicates the expected reward for a transition from state $s$ under action $a$, that is, $r(s,a) \doteq \mathbb{E}[R_{t+1} | S_t = s, A_t = a]$.

The value of a state $s$ when following a policy $\pi$, $v_\pi(s)$, is defined to be the expected sum of discounted rewards from that state: $v_\pi(s) \doteq \mathbb{E}_\pi\Big[\sum_{k=t+1}^T \gamma^{k - t - 1} R_{k} \Big | S_t = s \Big]$, where $\gamma$ is the discount factor. When the transition probability function $p$ and the reward function $r$ are known, we can compute $v_\pi(s)$ recursively by solving the system of equations below \cite{Bellman57}:
\begin{equation*}
v_\pi(s) = \sum\nolimits_{a} \pi(a|s) \Big [ r(s,a) + \gamma \sum\nolimits_{s'} p(s' | s, a) v_\pi(s')\Big ] .
\end{equation*}
These \ equations \ can \ also \ be \ written \ in \ matrix \ form \ with ${\bf v_\pi}$, ${\bf r} \in \mathbb{R}^{|\mathscr{S}|}$ and $P_\pi \in \mathbb{R}^{|\mathscr{S}| \times |\mathscr{S}|}$: 
\begin{eqnarray}
{\bf v_\pi} = {\bf r} + \gamma P_\pi {\bf v_\pi} = (I - \gamma P_\pi)\inv {\bf r}, \label{eq:value_function}
\end{eqnarray}
where $P_\pi$ is the state to state transition probability function induced by $\pi$, that is, $P_\pi(s, s') = \sum\nolimits_a \pi(a|s) p(s'|s, a)$.

Traditional model-based algorithms learn estimates of the matrix $P_\pi$ and~of~the~vector ${\bf r}$ and use them to estimate ${\bf v_\pi}$, for example by solving Equation~\ref{eq:value_function}. We use $\pHat$ and $\hat{\bf r}$ to denote empirical estimates of $P_\pi$ and ${\bf r}$.~Formally,
\begin{eqnarray}
\pHat(s'| s) = \frac{n(s, s')}{n(s)},  && \ \ \ \ \hat{\bf r}(s) = \frac{C(s,s')}{n(s)}, \label{eq:model_estimation}
\end{eqnarray}
where $\hat{\bf r}(i)$ denotes the $i$-th entry in the vector $\hat{\bf r}$, $n(s, s')$ is the number of times the transition $s \rightarrow s'$ was observed, $n(s) = \sum_{s' \in \mathscr{S}} n(s, s')$, and $C(s, s')$ is the sum of the rewards associated with the $n(s, s')$ transitions (we drop the action to simplify notation). However, model-based approaches are rarely successful in problems with large state spaces due to the difficulty in learning accurate models.

Because of the challenges in model learning, model-free solutions largely dominate the literature. In model-free RL, instead of estimating $P_\pi$ and ${\bf r}$, we estimate $v_\pi(s)$ directly from samples. We often use TD learning \cite{Sutton88} to update our estimates of $v_\pi(s)$, $\hat{v}(s)$, online:
\begin{eqnarray}
\hat{v}(S_t) \leftarrow \hat{v}(S_t) + \alpha \big[R_{t+1} + \gamma \hat{v}(S_{t+1}) - \hat{v}(S_t)\big],
\end{eqnarray}
where $\alpha$ is the step-size parameter. Generalization is required in problems with large state spaces, where it is unfeasible to learn an individual value for each state. We do so by parametrizing~$\hat{v}(s)$~with a set of weights $\theta$. We write, given the weights $\theta$, $\hat{v}(s; \theta) \approx v_\pi(s)$ and $\hat{q}(s, a; \theta) \approx q_\pi(s, a)$, where $q_\pi(s, a) = r(s, a) + \gamma \sum_{s'} p(s' | s,a) v_\pi(s')$. Model-free methods have performed well in problems with large state spaces, mainly due to the use of neural networks as function approximators (e.g., \citeauthor{Mnih15}~\citeyear{Mnih15}).

The ideas presented here are based on the successor representation (SR; \citeauthor{Dayan93} \citeyear{Dayan93}). The successor representation with respect to a policy $\pi$, $\Psi_\pi$, is defined as
\begin{equation*}
\Psi_\pi(s, s') = \mathbb{E}_{\pi, p} \Big[\sum_{t=0}^\infty \gamma^t \mathbb{I}\{S_t = s'\} \Big| S_0 = s \Big],
\end{equation*}
where we assume the sum is convergent with $\mathbb{I}$ denoting the indicator function. This expectation can actually be estimated from samples with TD learning:
\begin{eqnarray}
\hat{\Psi}(S_t, j) \!\!\!\! & \leftarrow & \!\!\!\!  \hat{\Psi}(S_t, j) + \eta \Big(\mathbb{I}\{S_t=j\} + \nonumber \\
                   & & \hspace{1.5cm} \gamma \hat{\Psi}(S_{t+1}, j) - \hat{\Psi}(S_t, j) \Big),
\end{eqnarray}
for all $j \in \mathscr{S}$ and $\eta$ denoting the step-size. The SR also corresponds to the Neumann series of $\gamma P_\pi$:
\begin{eqnarray}
\Psi_\pi = \sum_{t=0}^\infty (\gamma P_\pi)^t = (I - \gamma P_\pi)\inv.
\end{eqnarray}
Notice that the SR is part of the solution when computing a value function: ${\bf v_\pi} = \Psi_\pi {\bf r}$ (Equation~\ref{eq:value_function}). We use $\PsiHat$ to denote the SR computed through $\pHat$, the approximation of~$P_\pi$.

Successor features \cite{Barreto17} generalize the successor representation to the function approximation setting. We use the definition for the uncontrolled case.

\begin{definition}[Successor Features] For a given $0 \leq \gamma < 1$, policy $\pi$, and for a feature representation $\boldsymbol{\phi}(s) \in \mathbb{R}^d$, the successor features for a state $s$ are:
$$\boldsymbol{\psi}_{\pi}(s) = \mathbb{E}_{\pi, p} \Bigg[\sum_{t=0}^\infty \gamma^t \boldsymbol{\phi}(S_t) \Bigg | S_0 = s \Bigg].$$
\label{def:sf}
\end{definition}

Alternatively, in matrix form, we can write the successor features as $\Psi_\pi \! = \! \sum_{t=0}^\infty (\gamma P_\pi)^t \Phi \! = \! (I - \gamma P_\pi)\inv\Phi$, where $\Phi \in \mathbb{R}^{|\mathscr{S}| \times d}$ is a matrix encoding the feature representation of each state such that $\boldsymbol{\phi}(s) \in \mathbb{R}^d$. This definition reduces to the SR in the tabular case, where $\Phi \! = \! I$.

\section{$||\Psi(s)||$ as an Exploration Bonus}
\label{sec:theory}

Recent results have shown that the SR naturally captures the diffusion properties of the environment (e.g., \citeauthor{Machado18b} \citeyear{Machado18b}; \citeauthor{Wu19} \citeyear{Wu19}). 
Inspired by these results, in this section we argue that the SR can be explicitly used 
to promote exploration. We show that the norm of the SR, while it is being learned, behaves as an exploration bonus that rewards agents for visiting states it has visited less often. We first demonstrate this behavior empirically, in the tabular case. We then introduce the substochastic successor representation to provide some theoretical intuition that justifies this idea. In subsequent sections we show how these ideas carry over to the function approximation~setting.

\subsection{First Empirical Demonstration}\label{sec:model_free_tabular}

To demonstrate the usefulness of the norm of the SR as an exploration bonus, we first compare the performance of traditional Sarsa~\cite{Rummery94,Sutton98} to Sarsa+SR, which incorporates the norm of the SR as an exploration bonus in the Sarsa update. The update equation for Sarsa+SR is
\begin{eqnarray}
\hat{q}(S_t, A_t) \!\!\!\! & \leftarrow & \!\!\!\! \hat{q}(S_t, A_t) + \alpha \Big(R_t + \beta \frac{1}{||\hat{\Psi}(S_t)||_1} + \nonumber \\
                                        & & \hspace{0.8cm} \gamma \hat{q}(S_{t+1}, A_{t+1}) - \hat{q}(S_t, A_t)\Big),
\end{eqnarray}
where $\beta$ is a scaling factor and, at each time step $t$, $\hat{\Psi}(S_t, \cdot)$ is updated before $\hat{q}(S_t, A_t)$ as per Equation~4.

\begin{figure}[t]
    \centering
    \begin{subfigure}[b]{0.45\textwidth}
        \includegraphics[width=\textwidth]{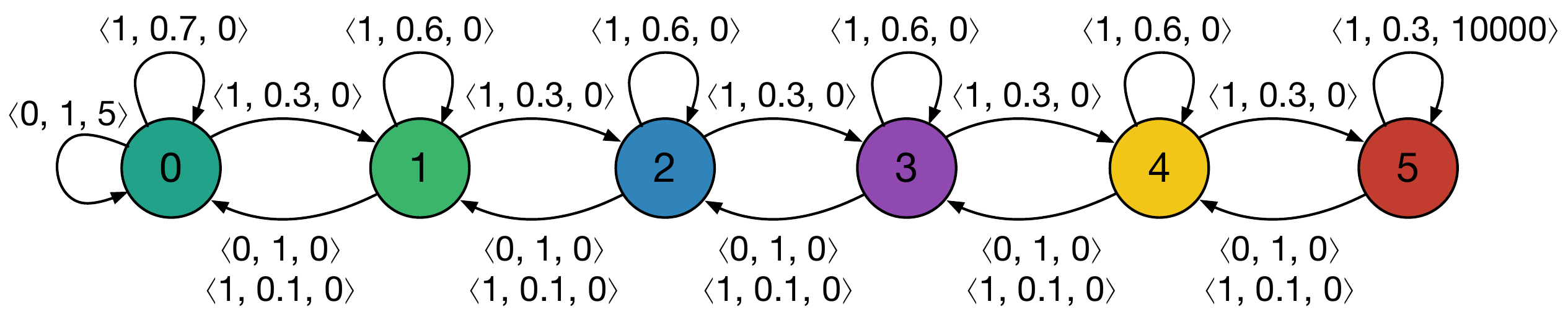}
        \caption{RiverSwim \linebreak}
    \end{subfigure}
    ~ 
      
    \begin{subfigure}[b]{0.42\textwidth}
        \includegraphics[width=\textwidth]{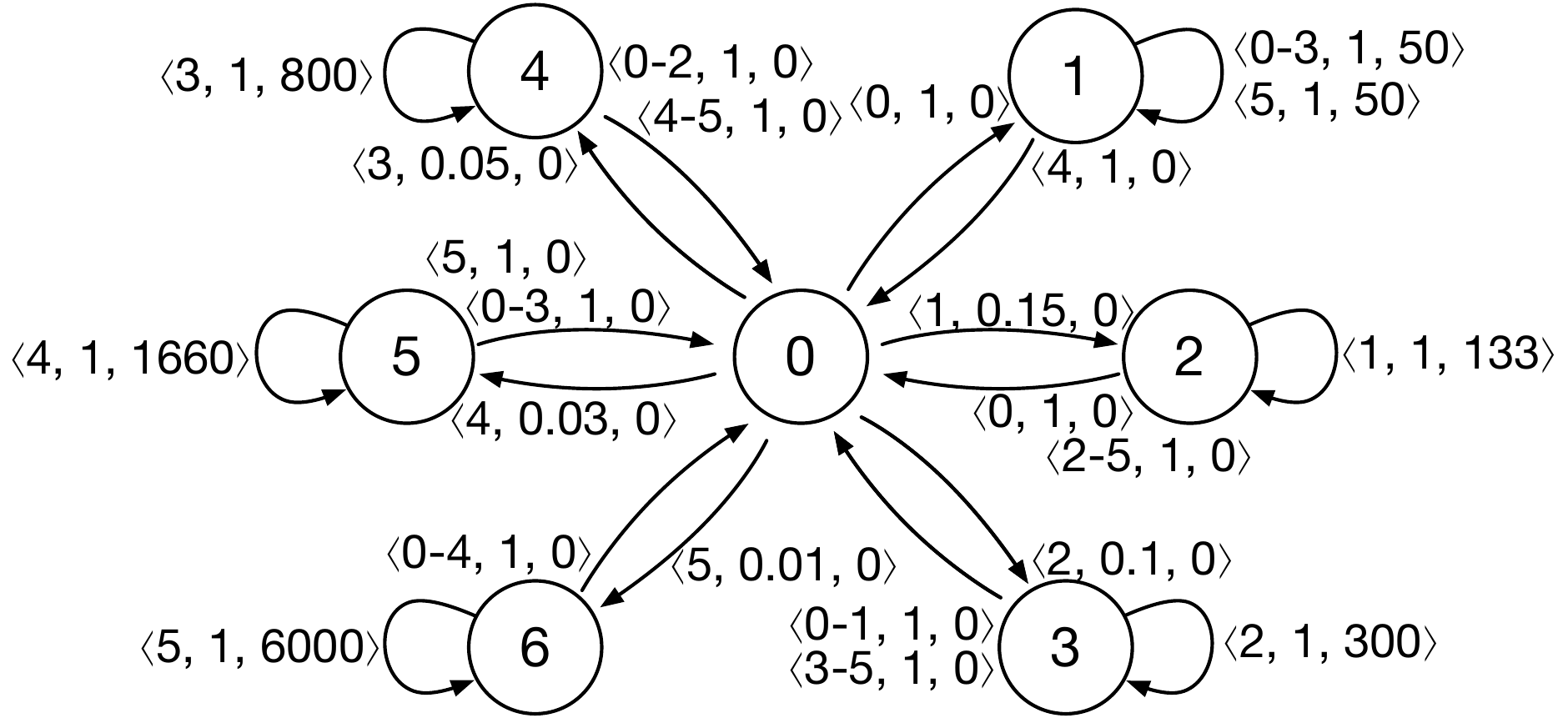}
        \caption{SixArms}
    \end{subfigure}
    \caption{Domains used in the tabular case. The tuples in each transition denote $\langle$action id, probability, reward$\rangle$. In \textsc{SixArms}, the agent starts in state~$0$. In \textsc{RiverSwim}, the agent starts in either state~$1$ or~$2$ with equal probability.} \label{fig:tabular_domains} 
\end{figure}

We evaluated this algorithm in \textsc{RiverSwim} and \textsc{SixArms}~\cite{Strehl08}, traditional domains in the PAC-MDP literature. In these domains, it is very likely that an agent will first observe a small reward generated in a state that is easy to get to. If the agent does not have a good exploration policy, it is likely to converge to a suboptimal behavior, never observing larger rewards available in states that are difficult to get to. See Figure~\ref{fig:tabular_domains} for more details. 

We compared the performance of Sarsa and Sarsa+SR for $5{,}000$ time steps when acting $\epsilon$-greedily to maximize the discounted return ($\gamma = 0.95$). For Sarsa+SR, we swept over different values of $\alpha$, $\eta$, $\gamma_{SR}$, $\beta$ and $\epsilon$, with $\alpha \in \{0.01, 0.05, 0.1, 0.25, 0.5\}$, $\eta \in \{0.01, 0.05, 0.1, 0.25, 0.5\}$, $\gamma_{SR} \in \{0.5, 0.8, 0.95, 0.99\}$, $\beta \in \{1, 10, 100, 1000, 10000\}$ and $\epsilon \in \{0.01, 0.05, 0.1\}$. For Sarsa, we swept over the parameters $\alpha$ and $\epsilon$. For fairness, we looked at a finer granularity for these parameters, with $\alpha \in i \times 0.005$ for $i$ ranging from $1$ to $100$, and with $\epsilon \in j \times 0.01$ for $j$ ranging from $1$ to $15$. Table~\ref{tab:parameters_tabular}, at the end of the paper, summarizes the parameter settings that led to the best results for each algorithm in \textsc{RiverSwim} and \textsc{SixArms}. The performance of each algorithm, averaged over $100$ runs, is available in Table~\ref{tab:results_tabular_sarsa}.

Our results show that the proposed exploration bonus has a profound impact in the algorithm's performance. Sarsa obtains an average return of approximately $25{,}000$ while Sarsa+SR obtains an approximate average return of $1.2$ million. Notice that, in \textsc{RiverSwim}, the reward that is ``easy to get'' has value~$5$, implying that, different from Sarsa+SR, Sarsa almost never explores the state space well enough. We observe the same trend in \textsc{SixArms}.

\subsection{Theoretical Justification}

It is difficult to characterize the behavior of our proposed exploration bonus because it is updated at each time step with TD learning. It is hard to analyze the behavior of estimates obtained with TD learning in the interim. Also, at its fixed point, for a fixed policy, the $\ell_1$-norm of the SR is $\sum \gamma^t 1 = 1/(1 - \gamma)$ for all states, preventing us from using the fixed point of the SR to theoretically analyze the behavior of this exploration bonus. In this section we introduce the substochastic successor representation (SSR) to provide some theoretical intuition, in the prediction case, of why the norm of the SR is a good exploration bonus. The SSR behaves similarly to the SR but it is simpler to analyze.

\begin{table}[t]
\centering
\caption{Comparison between Sarsa and Sarsa+SR. A 95\% confidence interval is reported between parentheses.}  \label{tab:results_tabular_sarsa}
\footnotesize
  \begin{tabular}{ l | r l | r l }
  & \multicolumn{2}{| c }{Sarsa} & \multicolumn{2}{| c }{Sarsa + SR} \\ \hline
  \textsc{RiverSwim}    &$24{,}770$   &\!\!\!($196$)   &\!\!\!$1{,}213{,}544$   &($540{,}454$)     \\ \hdashline[0.5pt/2pt]
  \textsc{SixArms}      &$247{,}977$  &\!\!\!($4{,}970$)  &\!\!\!$1{,}052{,}934$   &($2{,}311{,}617$) \\
  \end{tabular}
\end{table}

\begin{definition}[Substochastic Successor Representation] Let $\pTilde$ denote the substochastic matrix induced by the environment's dynamics and by the policy $\pi$ such that $\pTilde(s'| s) = \frac{n(s, s')}{n(s) + 1}.$ For a given $0 \leq \gamma < 1$, the substochastic successor representation, $\PsiTilde$, is defined as:
$$\PsiTilde = \sum_{t=0}^\infty \gamma^t \pTilde^t = (I - \gamma \pTilde)\inv.$$ \label{def:ssr}
\end{definition}

The SSR only differs from the empirical SR in its incorporation of an additional ``phantom'' transition from each state, making it underestimate the real SR. Through algebraic manipulation we show that the SSR allows us to recover an estimate of the visit counts, $n(s)$. This result provides some intuition of why the exploration bonus we propose performs so well, as exploration bonuses based on state visitation counts are known to generate proper exploration.

As aforementioned, the SSR behaves similarly to the SR. When computing the norm of the SR, while it is being learned with TD learning, it is as if a reward of 1 was observed at each time step.\footnote{In vector form, when estimating the SR with TD learning, the clause $\mathbb{I}\{S_t=j\}$, from Equation 4, is always true for one of the states, that is, an entry in the vector representing the SR. Thus, it is as if a reward of 1 was observed at each time step.} Thus, there is little variance in the target, with the predictions slowly approaching the true value of the SR. If pessimistically initialized, as traditionally done (i.e., initialized to zero when expecting positive rewards), the estimates of the SR approach the target from below. In this sense, the number of times a prediction has been updated in a given state is a good proxy to estimate how far this prediction is from its final target. From Definition 3.1 we can see that the SSR have similar properties. It underestimates the true target but slowly approaches it, converging to the true SR in the limit. The SSR simplifies the analysis by not taking bootstrapping into consideration.

The theorem below formalizes the idea that the norm of the SSR implicitly counts state visitation, shedding some light on the efficacy of the exploration bonus we propose.

\begin{theorem}~\label{theorem}
Let $n(s)$ denote the number of times state $s$ has been visited and let $\PsiTilde$ denote the substochastic successor representation as in Definition~\ref{def:ssr}. For a given $0 \leq \gamma < 1$,
\begin{eqnarray*}
\frac{\gamma}{n(s) \! + \! 1} \! - \! \frac{\gamma^2}{1 \! - \! \gamma} \leq (1 \! + \! \gamma) -||\PsiTilde(s)||_1 \leq \frac{\gamma}{n(s) \! + \! 1}.\\
\end{eqnarray*}
\end{theorem}

\begin{proof}[Proof of Theorem~\ref{theorem}] \let\qed\relax

Let $\pHat$ be the empirical transition matrix. We first rewrite $\pTilde$ in terms of $\pHat$:
\begin{eqnarray*}
\pTilde(s, s') \!\!\!\!\! &=& \!\!\!\!\!  \frac{n(s, s')}{n(s) + 1} \! = \! \frac{n(s)}{n(s) + 1} \frac{n(s, s')}{n(s)}\\
               \!\!\!\!\! &=& \!\!\!\!\! \frac{n(s)}{n(s) + 1} \pHat(s,s') \! = \! \Big(1 - \frac{1}{n(s) + 1}\Big) \pHat(s,s')
\end{eqnarray*}
This expression can also be written in matrix form: $\pTilde = (I - N)\pHat$, where $N \in \mathbb{R}^{|\mathscr{S}| \times |\mathscr{S}|}$ denotes the diagonal matrix of augmented inverse counts. Expanding $\PsiTilde$ we have:
\begin{eqnarray*}
\PsiTilde \ \ = \ \ \sum_{t=0}^\gamma (\gamma \pTilde)^t \ \ = \ \ I + \gamma \pTilde + \gamma^2 \pTilde^2 \PsiTilde.
\end{eqnarray*}

The top eigenvector of a stochastic matrix is the all-ones vector, ${\bf 1}$ \cite{meyn12markov}. 
Using this fact and the definition of $\pTilde$ with respect to $\pHat$ we have:
\begin{eqnarray}
\PsiTilde {\bf 1} &=& \big(I + \gamma(I - N)\pHat\big){\bf 1} + \gamma^2 \pTilde^2 \PsiTilde{\bf 1} \nonumber\\
          &=& (I + \gamma){\bf 1} - \gamma N {\bf 1} + \gamma^2 \pTilde^2 \PsiTilde{\bf 1} \label{eq:sr1}.
\end{eqnarray}
We can now bound the term $\gamma^2 \pTilde^2 \PsiTilde{\bf 1}$ using the fact that ${\bf 1}$ is also the top eigenvector of the successor representation and has eigenvalue $\frac{1}{1-\gamma}$~\cite{Machado18b}:
\begin{eqnarray*}
0 \leq \gamma^2 \pTilde^2 \PsiTilde{\bf 1} \leq \frac{\gamma^2}{1-\gamma}{\bf 1}.
\end{eqnarray*}
Plugging (\ref{eq:sr1}) into the definition of the SR we have (notice that $\Psi(s){\bf 1} = ||\Psi(s)||_1$):
\begin{eqnarray*}
\ (1 + \gamma){\bf 1}  - \PsiTilde{\bf 1} = \gamma N {\bf 1} - \gamma^2\pTilde^2\PsiTilde{\bf 1} \ \leq \ \gamma N {\bf 1}.
\end{eqnarray*}
When we also use the other bound on the quadratic term we conclude that, for any state $s$,
\begin{eqnarray*}
\frac{\gamma}{n(s) \! + \! 1} \! - \! \frac{\gamma^2}{1 \! - \! \gamma} \leq (1 \! + \! \gamma) \! - \! ||\PsiTilde(s)||_1 \leq \frac{\gamma}{n(s) \! + \! 1}. \ \qedsymbol
\end{eqnarray*} 
\end{proof}

\subsubsection{The relationship between the SSR and the SR.}
Theorem~1 shows that the SSR, obtained after a slight change to the SR, can be used to recover state visitation counts. The intuition behind this result is that the phantom transition, represented by the $+1$ in the denominator of the SSR, serves as a proxy for the uncertainty about that state by underestimating the SR. This is due to the fact that $\sum_{s'} \pTilde(s, s')$ gets closer to $1$ each time state $s$ is visited.

\begin{figure}[t]
    \centering
    \includegraphics[width=0.8\columnwidth]{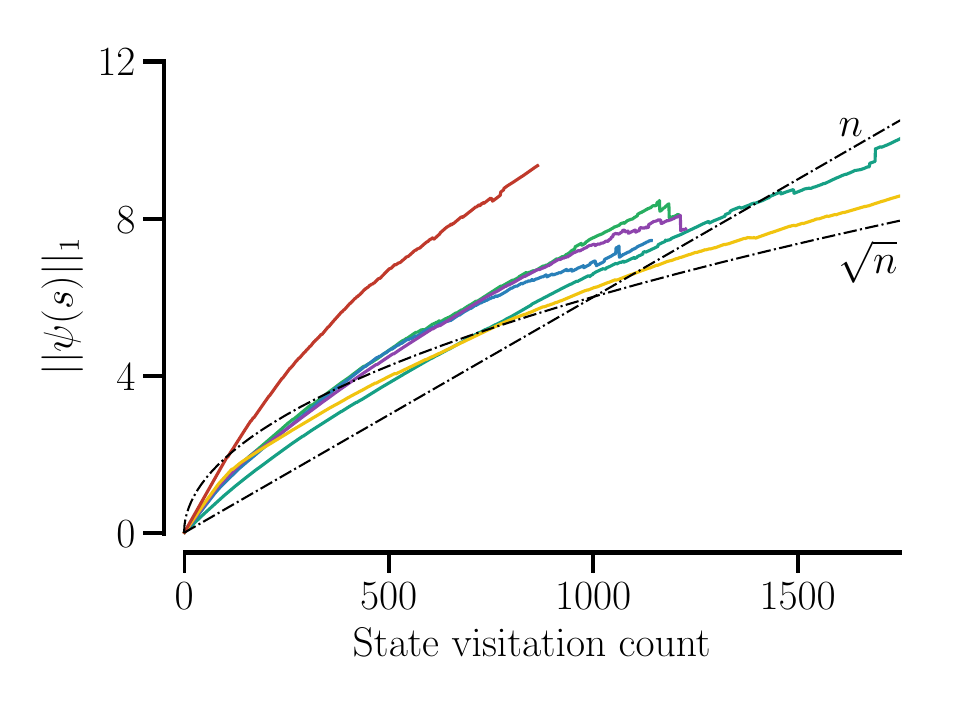}
    \caption{Empirical evaluation of the the norm of the SR as a function of state visitation count. Each curve denotes the evolution in one of the six states of \textsc{RiverSwim}. The colors match the colors of the states in Figure~\ref{fig:tabular_domains}a. Reference functions $f_1(n) = c_1 \sqrt{n}$ and $f_2(n) = c_2 n$ are depicted for comparison ($c_1 = 0.19$; $c_2 = 0.006$). See text for details.}\label{fig:empirical_count}
\end{figure}

This result also suggests that the exploration bonus we propose, $||\Psi(\cdot)||^{-1}$, behaves as a proper exploration bonus. It rewards the agent for visiting a state for the first times and it becomes less effective as the agent continues to visit that state. Hence, our approach does not change the optimal~policy because, in the limit, the norm of the SR (and the exploration bonus) converges to a constant value (see footnote~1):
\begin{eqnarray}\lim_{t \rightarrow \infty} ||\Psi(s)||_1 = \frac{1}{1-\gamma}.\end{eqnarray}

To validate the intuition that the SSR behaves similarly to the SR, we empirically evaluated the norm of the SR in the six states of RiverSwim as a function of the number of times those states were visited. Figure~\ref{fig:empirical_count} depicts the average result over the 100 runs used to generate the results in Table~\ref{tab:results_tabular_sarsa}. The plot shows that the norm of the SR does indeed grow as a function of state visitation counts. In this particular case, the norm of the SR grows at a rate between $1/n$, as suggested by the SSR, and $1/\sqrt{n}$, which is the rate at which TD often converges (Theorem 3.6; \citeauthor{Dalal18}~\citeyear{Dalal18}).

\subsubsection{Empirical validation of $||\tilde{\Psi}||_1$ as an exploration bonus.}

As a sanity check, we used the result in Theorem~1 to implement a simple model-based algorithm that penalizes the agent for visiting commonly visited states with the exploration bonus ${\bf r_{\mbox{\scriptsize int}}} = -||\PsiTilde(s)||_1$. Our agent maximizes $r(s,a) + \beta r_{\mbox{\scriptsize int}}(s)$, where $\beta$ is a scaling parameter. The shift $(1 + \gamma)$ in the theorem has no effect in the agent's policy because it is the same across all states. In this algorithm the agent updates its transition model and reward model with Equation~\ref{eq:model_estimation} and its SSR estimate as in Definition~\ref{def:ssr}.

Table~\ref{tab:results_tabular} depicts the performance of this algorithm, dubbed ESSR, as well as the performance of some algorithms with polynomial sample-complexity bounds. The goal with this evaluation is not to outperform these algorithms, but to evaluate how well ESSR performs when compared to algorithms that explicitly keep visitation counts to promote exploration. ESSR performs as well as \textsc{R-Max}~\cite{Brafman02} and E$^3$~\cite{Kearns02} on \textsc{RiverSwim} and it outperforms these algorithms on \textsc{SixArms}; while MBIE~\cite{Strehl08}, which explicitly estimates confidence intervals over the expected return in each state, outperforms ESSR in these domains. These results clearly show that ESSR performs, on average, similarly to other algorithms with PAC-MDP guarantees, suggesting that the norm of the SSR is a promising exploration bonus.\footnote{The code used to generate all results in this section is available at:
\url{https://github.com/mcmachado/count_based_exploration_sr/tree/master/tabular}.}

\begin{table}[t]
\centering
\caption{Comparison between ESSR, \textsc{R-Max}, \textsc{E}$^3$, and \textsc{MBIE}. The numbers reported for \textsc{R-Max}, \textsc{E}$^3$, and \textsc{MBIE} were extracted from the histograms presented by \citeauthor{Strehl08}~(\citeyear{Strehl08}). ESSR's performance is the average over $100$ runs. A $95\%$ confidence interval is reported between parentheses. All numbers are reported in millions (i.e., $\times 10^6$).}  \label{tab:results_tabular}
\small
  \begin{tabular}{ l | c | c | c| r l }
  & \textsc{E}$^3$ &\textsc{R-Max} &\textsc{MBIE} &\multicolumn{2}{| c }{\textsc{ESSR}}\\ \hline
  \textsc{RiverSwim}    &$3.0$  &$3.0$    &$3.3$    &$3.1$ &($0.06$)  \\ \hdashline[0.5pt/2pt]
  \textsc{SixArms}      &$1.8$   &$2.8$    &$9.3$    &$7.3$ &($1.2$)  \\
  \end{tabular}
\end{table}


\section{Counting Feature Activations with the SR}
\label{sec:deep_sr}

In large environments, where enumerating all states is not an option, directly using Sarsa+SR as described in the previous section is not viable. However, one of the reasons the results in the previous section are interesting is the fact that there is a natural extension of the SR to non-tabular settings, the successor features (Definition~2.1), and the fact that we can immediately use norms in the function approximation setting. In this section we show how one can extend the idea of using the norm of the SR as an exploration bonus to the function approximation setting, something one cannot easily do if relying on explicit state visitation counts. Because deep RL approaches often lead to state-of-the-art performance while also learning a representation from high-dimensional sensory inputs, in this section we introduce a deep RL algorithm that incorporates the ideas introduced. Our algorithm was also inspired by recent work that has shown that successor features can be learned jointly with the feature representation itself \cite{Kulkarni16,Machado18b}.

\begin{figure}[t]
    \centering
    \includegraphics[width=0.92\columnwidth]{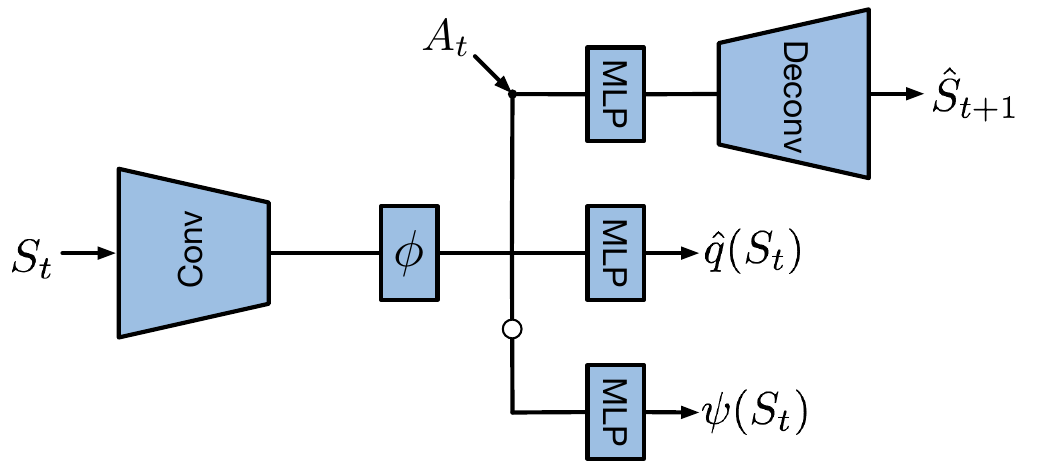}
    \caption{Neural network architecture used by our algorithm when learning to play Atari 2600 games.}\label{fig:network}
\end{figure}

An overview of the neural network we used to learn the agent's value function while also learning the feature representation and the SR is depicted in Figure~\ref{fig:network}. The layers used to compute the state-action value function, $\hat{q}(S_t, \cdot)$, are structured as in DQN~\cite{Mnih15}, with the number of parameters (i.e., filter sizes, stride, and number of nodes) matching \citeauthor{Oh15}'s~(\citeyear{Oh15}) architecture, which is known to succeed in the auxiliary task detailed below of predicting the agent's next observation. We call the part of our architecture that predicts $\hat{q}(S_t, \cdot)$ DQN$_e^{\scriptsize \mbox{MMC}}$. It is trained to minimize
\begin{eqnarray*}
\mathcal{L}_{\scriptsize \mbox{TD}} = \mathbb{E}\Big[\big((1 - \tau) \delta(s,a) + \tau \delta_{\scriptsize \mbox{MC}}(s,a)\big)^2\Big],
\end{eqnarray*}
with $\delta(s,a)$ and $\delta_{\scriptsize \mbox{MC}}(s,a)$ being defined as 
\begin{eqnarray*}
\delta(s,a) &=& R_t + \!\beta r_{\mbox{\scriptsize int}}(s; \theta^{-}) + \\ & & \ \ \ \ \ \ \ \ \ \ \ \ \ \ \ \ \ \ \gamma \max_{a'} q(s', a'; \theta^{-}) - q(s, a; \theta),\\
\delta_{\scriptsize \mbox{MC}}(s,a) \!\!\!\!\! &=& \!\!\!\!\! \sum_{t=0}^\infty \gamma^t \Big(r(S_t, A_t) \! + \! \beta r_{\mbox{\scriptsize int}}(S_t; \theta^{-})\Big) \! - \! q(s, a; \theta).
\end{eqnarray*}

This loss is known as the mixed Monte-Carlo return (MMC) and it has been used in the past by the algorithms that achieved succesful exploration in deep reinforcement learning~\cite{Bellemare16,Ostrovski17}. The distinction between $\theta$ and $\theta^{-}$ is standard in the field, with $\theta^{-}$ denoting the parameters of the target network, which is updated less often for stability purposes~\cite{Mnih15}. As before, we use $r_{\mbox{\scriptsize int}}$ to denote the exploration bonus obtained from the successor features of the internal representation, $\phi$, which will be defined below. Moreover, to ensure all features are in the same range, we normalize the feature vector so that $||\phi(\cdot)||_1 = 1$. In Figure~\ref{fig:network} we highlight with $\phi$ the layer in which we normalize its output. Notice that the features are always non-negative due to the~use~of~ReLU~gates.

\begin{table*}[t]
\centering
\caption{Performance of the proposed algorithm, DQN$_e^{\scriptsize \mbox{MMC}}$+SR, compared to various agents on the ``hard exploration'' subset of Atari 2600 games. The DQN results reported are from \citeauthor{Machado18a}~(\citeyear{Machado18a}) while the \textsc{DQN$^{\scriptsize \mbox{MMC}}_{\mbox{\scriptsize{CTS}}}$}, \textsc{DQN$^{\scriptsize \mbox{MMC}}_{\mbox{\scriptsize{PixelCNN}}}$} and RND results were obtained through personal communication with the authors of the corresponding papers. \citeauthor{Burda19} did not evaluate RND in \textsc{Freeway}. When available, standard deviation is reported between parentheses. See text for details.}  \label{tab:results_ale}
\footnotesize
  \begin{tabular}{ l | r l | r l | l | l | r l |  r l}
  &\multicolumn{2}{| c |}{DQN} &\multicolumn{2}{| c | }{DQN$_e^{\scriptsize \mbox{MMC}}$} &\textsc{DQN$^{\scriptsize \mbox{MMC}}_{\mbox{\scriptsize{CTS}}}$} &\textsc{DQN$^{\scriptsize \mbox{MMC}}_{\mbox{\scriptsize{PixelCNN}}}$} &\multicolumn{2}{| c}{RND} &\multicolumn{2}{| c }{DQN$_e^{\scriptsize \mbox{MMC}}$+SR}\\ \hline
  \textsc{Freeway}                &32.4   &(0.3)     & 29.5   &(0.1)   & \ \ \ \ 29.2   & \ \ \ \ \ \ \ 29.4    &-       &-               &29.4   &(0.1)   \\ \hdashline[0.5pt/2pt]
  \textsc{Gravitar}               &118.5  &(22.0)    & 1078.3 &(254.1) & \ \ \ \ 199.8  & \ \ \ \ \ \ \ 275.4   &790.0   &(122.9)         &457.4  &(120.3) \\ \hdashline[0.5pt/2pt]
  \textsc{Mont. Rev.}             &0.0    &(0.0)     & 0.0    &(0.0)   & \ \ \ \ 2941.9 & \ \ \ \ \ \ \ 1671.7  &524.8   &(314.0)         &1395.4 &(1121.8) \\ \hdashline[0.5pt/2pt]
  \textsc{Private Eye}            &1447.4 &(2,567.9) & 113.4  &(42.3)  & \ \ \ \ 32.8   & \ \ \ \ \ \ \ 14386.0 &61.3    &(53.7)          &104.4  &(50.4)   \\ \hdashline[0.5pt/2pt]
  \textsc{Solaris}                &783.4  &(55.3)    & 2244.6 &(378.8) & \ \ \ \ 1147.1 & \ \ \ \ \ \ \ 2279.4  &1270.3  &(291.0)         &1890.1 &(163.1) \\ \hdashline[0.5pt/2pt]
  \textsc{Venture}                &4.4    &(5.4)     & 1220.1 &(51.0)  & \ \ \ \ 0.0    & \ \ \ \ \ \ \ 856.2   &953.7   &(167.3)         &1348.5 &(56.5) \\
  \end{tabular} 
\end{table*}

The successor features, $\psi(S_t)$, at the bottom of the diagram, are obtained by minimizing~the~loss
\begin{eqnarray*}
\mathcal{L}_{\scriptsize\mbox{SR}} = \mathbb{E} \Big[ \big( \phi(S_t; \theta^-) + \gamma \psi(S_{t+1}; \theta^-) - \psi(S_t; \theta) \big)^2 \Big].
\end{eqnarray*}
Zero is a fixed point for the SR, which is particularly concerning in settings with sparse rewards. The agent might end up learning to set $\phi(\cdot) = \vec{0}$ to achieve zero loss. We address this problem by not propagating $\nabla \mathcal{L}_{\scriptsize \mbox{SR}}$ to $\phi$ (this is depicted in Figure~\ref{fig:network} as an open circle stopping the gradient); and by creating an auxiliary task~\cite{Jaderberg17} to encourage a representation to be learned before a non-zero reward is observed. As \citeauthor{Machado18b}~(\citeyear{Machado18b}), we use the auxiliary task of predicting the next observation, learned with the architecture proposed by \citeauthor{Oh15}~(\citeyear{Oh15}), which is depicted as the top layers in Figure~\ref{fig:network}. The loss we minimize for this last part of the network is $$\mathcal{L}_{\scriptsize \mbox{Recons}} = \big(\hat{S}_{t+1} - S_{t+1}\big)^2.$$

The overall loss minimized by the network is $$\mathcal{L} = w_{\scriptsize\mbox{TD}} \mathcal{L}_{\scriptsize \mbox{TD}} + w_{\scriptsize\mbox{SR}} \mathcal{L}_{\scriptsize \mbox{SR}} + w_{\scriptsize\mbox{Recons}} \mathcal{L}_{\scriptsize \mbox{Recons}}.$$

The last step in describing our algorithm is to define $r_{\mbox{\scriptsize int}}(S_t; \theta^{-})$, the intrinsic reward we use to encourage exploration. As in Sarsa+SR, we choose the exploration bonus to be the inverse of the $\ell_1$-norm of the vector of successor features of the current state. That is,
\begin{eqnarray*}
r_{\mbox{\scriptsize int}}(S_t; \theta^{-}) = \frac{1}{||\psi(S_t; \theta^{-})||_1},
\end{eqnarray*}
where $\psi(S_t; \theta^{-})$ denotes the successor features of state $S_t$ parametrized by $\theta^{-}$. The exploration bonus comes from the same intuition presented in the previous section 
(we observed in preliminary experiments not discussed here that DQN performs better when dealing with positive rewards). 

\begin{figure*}[t]
    \centering
    \includegraphics[width=0.7\textwidth]{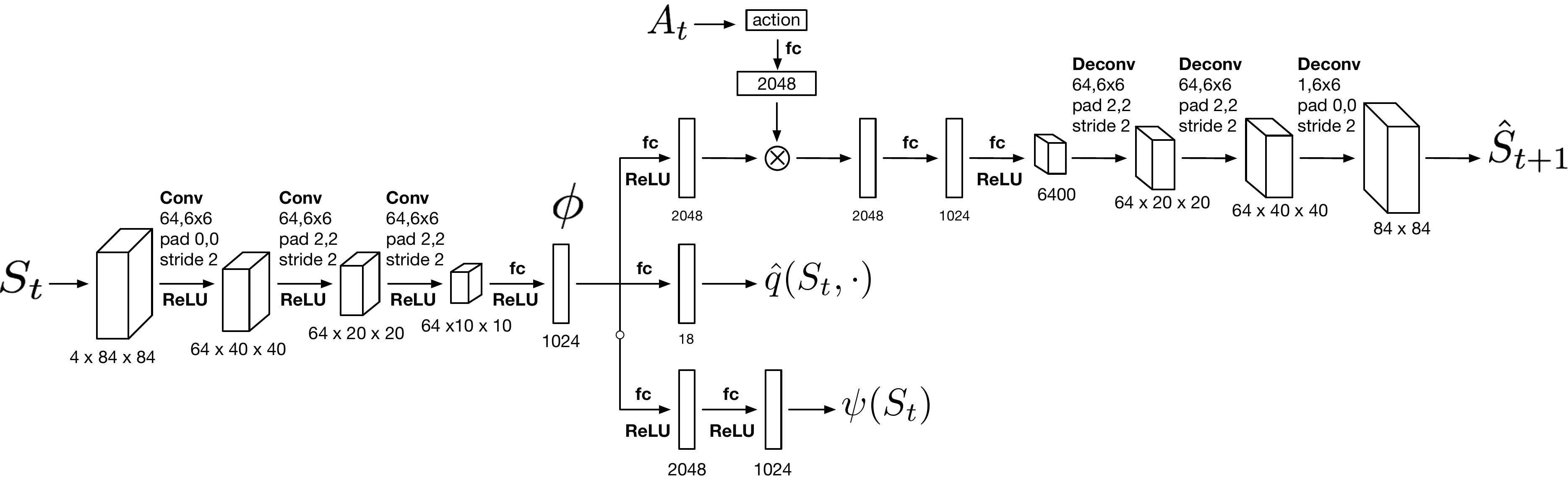}
    \caption{Neural network architecture used by our algorithm when learning to play Atari 2600 games.}\label{fig:detailed_network}
\end{figure*}

A complete description of the network architecture is available in Figure~\ref{fig:detailed_network}, which is at the end of the paper to allow the reader to first focus on the main concepts of the proposed idea. We initialize our network the same way \citeauthor{Oh15}~(\citeyear{Oh15}) does. We use Xavier initialization~\cite{Glorot10} in all layers except the fully connected layers around the element-wise multiplication denoted by $\otimes$, which are initialized uniformly with values between $-0.1$ and $0.1$. 

\section{Evaluation of Exploration in Deep RL}
\label{sec:evaluation}

We evaluated our algorithm on the Arcade Learning Environment~\cite{Bellemare13}. Following \citeauthor{Bellemare16}'s~(\citeyear{Bellemare16}) taxonomy, we focused on the Atari 2600 games with sparse rewards that pose hard exploration problems. They are: \textsc{Freeway, Gravitar, Montezuma's Revenge, Private Eye, Solaris}, and \textsc{Venture}.\footnote{The code used to generate the reported results is available at:\\
\url{https://github.com/mcmachado/count_based_exploration_sr/tree/master/function_approximation}.}

We used the evaluation protocol proposed by~\citeauthor{Machado18a}~(\citeyear{Machado18a}). 
The reported results are the average over 10 seeds after 100 million frames. We evaluated our agents in the stochastic setting (sticky actions, $\varsigma = 0.25$) using a frame skip of $5$ with the full action set ($|\mathscr{A}| = 18$). The agent learns from raw pixels i.e., it~uses~the~game~screen~as~input.

Our results were obtained with the algorithm described in Section~\ref{sec:deep_sr}. We set $\beta=0.05$ after a rough sweep over values in the game \textsc{Montezuma's Revenge}. We annealed $\epsilon$ in DQN's $\epsilon$-greedy exploration over the first million steps, starting at $1.0$ and stopping at $0.1$ as done by \citeauthor{Bellemare16}~(\citeyear{Bellemare16}). We trained the network with RMSprop with a step-size of $0.00025$, an $\epsilon$ value of $0.01$, and a decay~ of~$0.95$, which are the standard parameters for training DQN~\cite{Mnih15}. The discount factor, $\gamma$, is set to $0.99$, and $w_{\scriptsize\mbox{TD}} = 1$, $w_{\scriptsize\mbox{SR}} = 1000$, $w_{\scriptsize\mbox{Recons}} = 0.001$. The weights  $w_{\scriptsize\mbox{TD}}$, $w_{\scriptsize\mbox{SR}}$, and $w_{\scriptsize\mbox{Recons}}$ were set so that the loss functions would be roughly at the same scale. All other parameters are the same as those used by \citeauthor{Mnih15}~(\citeyear{Mnih15}) and \citeauthor{Oh15}~(\citeyear{Oh15}).

\subsection{Overall Performance and Baselines}

Table~\ref{tab:results_ale} summarizes the results after 100 million frames. The performance of other algorithms is also provided for reference. Notice we are reporting learning performance for all algorithms instead of the maximum scores achieved by the algorithm. We use the superscript $^{\scriptsize \mbox{MMC}}$ to distinguish between the algorithms that use MMC from those that do not. When comparing our algorithm, DQN$_e^{\scriptsize \mbox{MMC}}$+SR, to DQN we can see how much our approach improves over the most traditional baseline. By comparing our algorithm's performance to \textsc{DQN$^{\scriptsize \mbox{MMC}}_{\mbox{\scriptsize{CTS}}}$}~\cite{Bellemare16} and \textsc{DQN$^{\scriptsize \mbox{MMC}}_{\mbox{\scriptsize{PixelCNN}}}$}~\cite{Ostrovski17} we compare our algorithm to established baselines for exploration that are closer to our method. By comparing our algorithm's performance to Random Network Distillation~(RND; \citeauthor{Burda19} \citeyear{Burda19}) we compare our algorithm to the most recent paper in the field with state-of-the-art performance. Finally, we also evaluate the impact of the proposed exploration bonus by comparing our algorithm to DQN$_e^{\scriptsize \mbox{MMC}}$, which uses the same number of parameters our network uses (i.e., filter sizes, stride, and number of nodes), but without the additional modules (next state prediction and successor representation) and without the intrinsic reward bonus. We do this by setting $w_{\scriptsize\mbox{SR}} = w_{\scriptsize\mbox{Recons}} = \beta = 0$.

We can clearly see that our algorithm achieves scores much higher than those achieved by DQN, which struggles in games that pose hard exploration problems.  When comparing our algorithm to \textsc{DQN$^{\scriptsize \mbox{MMC}}_{\mbox{\scriptsize{CTS}}}$} and \textsc{DQN$^{\scriptsize \mbox{MMC}}_{\mbox{\scriptsize{PixelCNN}}}$} we observe that, on average, DQN$_e^{\scriptsize \mbox{MMC}}$+SR at least matches the performance of these algorithms while being simpler by not requiring a density model. Instead, our algorithm requires the SR, which is domain-independent as it is already defined for every problem since it is a component of the value function estimates, as discussed in Section~\ref{sec:background}. Finally, DQN$_e^{\scriptsize \mbox{MMC}}$+SR also outperforms RND~\cite{Burda19} when it is trained for 100 million frames.\footnote{DQN$_e^{\scriptsize \mbox{MMC}}$+SR outperforms \textsc{DQN$^{\scriptsize \mbox{MMC}}_{\mbox{\scriptsize{CTS}}}$} in five out of six games, it outperforms RND in four out of five games, and its performance is comparable to \textsc{DQN$^{\scriptsize \mbox{MMC}}_{\mbox{\scriptsize{PixelCNN}}}$}'s performance.} Importantly, RND, when trained for 2 billion frames, is currently considered to be the state-of-the-art approach for exploration in Atari 2600 games. Recently \citeauthor{Taiga19} (\citeyear{Taiga19}) evaluated several exploration algorithms, including~those~we~use as baselines, and they have shown that, in these games, their performance at 100 million frames is predictive of their performance at one billion frames.

Finally, the comparison between DQN$_e^{\scriptsize \mbox{MMC}}$+SR and DQN$_e^{\scriptsize \mbox{MMC}}$ shows that the provided exploration bonus has a big impact in the game \textsc{Montezuma's Revenge}, which is probably known as the hardest game among those we used in our evaluation, and the only game where agents do not learn how to achieve scores greater than zero with random exploration. Interestingly, the change in architecture and the use of MMC leads to a big improvement in games such as \textsc{Gravitar} and \textsc{Venture}, which we cannot fully explain. However, because the change in architecture does not have any effect in \textsc{Montezuma's Revenge}, it seems that the proposed exploration bonus is essential in games with very sparse rewards.

\subsection{Evaluating the Impact of the Auxiliary Task}

While the results depicted in Table~3 allow us to see the benefit of using an exploration bonus derived from the SR, they do not inform us about the impact of the auxiliary task in the results. The experiments in this section aim at addressing this issue. We focus on \textsc{Montezuma's Revenge} because it is the game where the problem of exploration is maximized, with most algorithms not being able to do anything without an exploration bonus.

The first question we asked was whether the \emph{auxiliary task was necessary} in our algorithm. We evaluated this by dropping the reconstruction module from the network to test whether the initial random noise generated by the SR is enough to drive representation learning. It is not. When dropping the auxiliary task, the average performance of this baseline over 4 seeds in \textsc{Mont. Revenge} after $100$ million frames was $100$ points ($\sigma^2$ = $200$; min:~$0$, max:~$400$). As comparison, our algorithm obtains $1395.4$ points ($\sigma^2$ = $1121.8$, min: $400$, max: $2500$). These results suggest that auxiliary tasks are necessary for our method to perform well.

We also evaluated whether the \emph{auxiliary task was sufficient} to generate the results we observed. To do so we dropped the SR module and set $\beta=0.0$ to evaluate whether our exploration bonus was actually improving the agent's performance or whether the auxiliary task was doing it. The exploration bonus seems to be essential. When dropping the exploration bonus and the SR module, the average performance of this baseline over 4 seeds in \textsc{Montezuma's Revenge} after $100$ million frames was $398.5$ points ($\sigma^2$ = $230.1$; min:~$0$, max:~$400$). Again, clearly, the auxiliary task is not a sufficient condition for the performance we report. The reported results use the same parameters as before. 

Thus, while it is hard to completely disentangle the impact of the different components of DQN$_e^{\scriptsize \mbox{MMC}}$+SR (e.g., the exploration bonus, learning the SR, the auxiliary task), the comparisons in Table~\ref{tab:results_ale} and the results in this section suggest that the exploration bonus we introduced is essential for our approach to achieve state-of-the-art performance.

\subsection{Evaluating the Impact of Using Different P-Norms}

To further understand the different nuances behind the idea that the norm of the successor representation can be used to generate an exploration bonus, we also asked the question of whether this is true only for the $\ell_1$-norm of the SR. It is often said, for example, that the $\ell_2$-norm is smoother, and thus might be more amenable to the training of neural networks.

For the function approximation case, DQN$_e^{\scriptsize \mbox{MMC}}$+SR when using the $\ell_2$-norm has a performance comparable to DQN$_e^{\scriptsize \mbox{MMC}}$+SR when using the $\ell_1$-norm of the SR to generate its exploration bonus ($\phi$ is normalized with the respective norm). The actual performance of both approaches in the Atari 2600 games we used in the previous experiment is available in  Table~\ref{tab:mismatch_atari}. We followed the same evaluation protocol described before, averaging the performance of DQN$_e^{\scriptsize \mbox{MMC}}$+SR with the $\ell_2$-norm over 10 runs. The parameter $\beta$ is the only parameter not shared by both algorithms. While $\beta=0.025$ when using the $\ell_2$-norm of the SR, $\beta=0.05$ when using the $\ell_1$-norm of the SR. 

We also revisited the results presented in Section~\ref{sec:model_free_tabular} to evaluate, in the tabular case, the impact of the different norms in Sarsa+SR. We swept over all the parameters, as previously described. The results reported for Sarsa+SR when using the $\ell_2$-norm of the SR are the average over 100 runs. The actual numbers are available in Table~\ref{tab:mismatch_tabular}. As before, it seems that it does not make much difference which norm of the SR we use ($\ell_1$-norm or the $\ell_2$-norm). The fact that these results are so close might suggest that the idea of using the norm of the SR for exploration is quite general, with the $p$-norm of the SR being effective for more than one value of $p$. Recall that $||x||_1 \leq \sqrt{2}||x||_2$ for any finite vector $x$.

\begin{table}[t]
\centering
\caption{Performance of the proposed algorithm, DQN$_e^{\scriptsize \mbox{MMC}}$+SR, when using the $\ell_1$-norm and $\ell_2$-norm of the SR to generate the exploration bonus. Standard~deviation is reported between parentheses. See text for~details.}  \label{tab:mismatch_atari}
\footnotesize
  \begin{tabular}{ l | r l | r l}
  &\multicolumn{2}{| c }{$\ell_1$-norm} &\multicolumn{2}{| c }{$\ell_2$-norm}\\ \hline
  \textsc{Freeway}                & 29.4    &(0.1)      &29.5   &(0.1)   \\ \hdashline[0.5pt/2pt]
  \textsc{Gravitar}               & 457.4   &(120.3)    &430.3  &(109.4) \\ \hdashline[0.5pt/2pt]
  \textsc{Mont. Rev.}             & 1395.4  &(1121.8)   &1778.6 &(903.6) \\ \hdashline[0.5pt/2pt]
  \textsc{Private Eye}            & 104.4   &(50.4)     &99.1   &(1.8)   \\ \hdashline[0.5pt/2pt]
  \textsc{Solaris}                & 1890.1  &(163.1)    &2155.7 &(398.3) \\ \hdashline[0.5pt/2pt]
  \textsc{Venture}                & 1348.5  &(56.5)     &1241.8 &(236.0) \\
  \end{tabular} 
\end{table}

\begin{table}[t]
\centering
\caption{Performance of Sarsa+SR when using the $\ell_1$-norm and $\ell_2$-norm of the SR to generate the exploration bonus. A 95\% confidence interval is reported between parentheses.}  \label{tab:mismatch_tabular}
\footnotesize
  \begin{tabular}{ l | r l | r l}
  &\multicolumn{2}{| c }{$\ell_1$-norm} &\multicolumn{2}{| c }{$\ell_2$-norm}\\ \hline
  \!\!\! \textsc{RiverSwim}\!\!  & \!\!\! $1{,}213{,}544$     &\!\!\!\!\!\!($540{,}454$)\!\!      & \!\!\! $1{,}192{,}052$   &\!\!\!\!\!\!($507{,}179$)   \\ \hdashline[0.5pt/2pt]
	  \!\!\! \textsc{SixArms}  \!\!  & \!\!\! $1{,}052{,}934$     &\!\!\!\!\!\!($2{,}311{,}617$)\!\!   & \!\!\! $819{,}927$       &\!\!\!\!\!\!($2{,}132{,}003$) \\
  \end{tabular} 
\end{table}

\section{Related Work}

\begin{table*}[t]
\centering
\caption{Parameter settings that led to the reported performance in \textsc{RiverSwim} and \textsc{SixArms}.}\label{tab:parameters_tabular}
\small{
  \begin{tabular}{ l | c | c | c | c | c | c | c | c | c | c }
                                                        &\multicolumn{5}{| c }{\textsc{RiverSwim}} &\multicolumn{5}{| c }{\textsc{SixArms}} \\
    Algorithm                                           &$\alpha$  &$\eta$  &$\gamma_{\mbox{\scriptsize SR}}$  &$\beta$      & $\epsilon$  &$\alpha$  &$\eta$  &$\gamma_{\mbox{\scriptsize SR}}$  &$\beta$      & $\epsilon$ \\ \hline
    Sarsa                                               &$0.005$   & -      & -                    & -           &$0.01$  &$0.465$    & -      & -                    & -           &$0.03$       \\ \hdashline[0.5pt/2pt]
    Sarsa+SR (w/ $\ell_1$-norm)      &$0.25$     &$0.01$   &$0.95$              &$100$        &$0.1$   &$0.1$     &$0.01$  &$0.99$                 &$100$   &$0.01$       \\ \hdashline[0.5pt/2pt]
    Sarsa+SR (w/ $\ell_2$-norm)      &$0.25$     &$0.01$   &$0.99$              &$100$        &$0.1$   &$0.1$     &$0.01$  &$0.99$                 &$10$    &$0.01$
  \end{tabular}
}
\end{table*}

There are multiple algorithms in the tabular, model-based case, with guarantees about their performance in terms of regret bounds (e.g., \citeauthor{Osband16b} \citeyear{Osband16b}) or sample-complexity (e.g., \citeauthor{Brafman02} \citeyear{Brafman02}; \citeauthor{Kearns02} \citeyear{Kearns02}; \citeauthor{Strehl08} \citeyear{Strehl08}). \textsc{RiverSwim} and \textsc{SixArms} are domains traditionally used when evaluating these algorithms. In this paper we introduced a model-free algorithm that performs particularly well in these domains. We also introduced a model-based algorithm that performs as well as some of these algorithms with theoretical guarantees. Among these algorithms, \textsc{R-Max} is the closest approach to ours. As \textsc{R-Max}, the algorithm presented in Section~\ref{sec:theory} augments the state-space with an imaginary state and encourages the agent to visit that state, implicitly reducing the algorithm's uncertainty. However, \textsc{R-Max} deletes the transition to this imaginary state once a state has been visited a given number of times. Ours, on the other hand, lets the probability of visiting this imaginary state vanish with additional~visitations. Importantly, notice that it is not clear how to apply traditional algorithms such as \textsc{R-Max} and E$^3$ to large domains where function approximation~is~required.

Conversely, there are not many model-free approaches with proven sample-complexity bounds (e.g., \citeauthor{Strehl06} \citeyear{Strehl06}), but there are multiple model-free algorithms for exploration that actually work in large domains~(e.g., \citeauthor{Bellemare16} \citeyear{Bellemare16}; \citeauthor{Ostrovski17} \citeyear{Ostrovski17}; \citeauthor{Plappert18} \citeyear{Plappert18}; \citeauthor{Burda19} \citeyear{Burda19}). Among these algorithms, the use of pseudo-counts through density models is the closest to ours~\cite{Bellemare16,Ostrovski17}. Inspired by those papers we used the mixed Monte-Carlo return as a target in the update rule. In Section~\ref{sec:evaluation} we showed that our algorithm at least matches these approaches while being simpler by not requiring a density model. Importantly, \citeauthor{Martin17}~(\citeyear{Martin17}) had already shown that counting activations of fixed, handcrafted features in Atari 2600 games leads to good exploration behavior. Nevertheless, by using the SR we are not only counting \emph{learned} features but we are also implicitly capturing the induced transition dynamics.

\section{Conclusion}

RL algorithms tend to have high sample complexity, which often prevents them from being used in the real-world. Poor exploration strategies is one of the reasons for this high sample-complexity. Despite all of its shortcomings, uniform random exploration is, to date, the most commonly used approach for exploration. This is mainly due to the fact that most approaches for tackling the exploration problem still rely on domain-specific knowledge (e.g., density models, handcrafted features), or on having an agent learn a perfect model of the environment. In this paper we introduced a general method for exploration in RL that implicitly counts state (or feature) visitation in order to guide the exploration process. It is compatible with representation learning and the idea can also be adapted to be applied to large domains.

This result opens up multiple possibilities for future work. Based on the results presented in Section~\ref{sec:theory}, for example, we conjecture that the substochastic successor representation can be actually used to generate algorithms with PAC-MDP bounds (e.g., one could replace explicit state visitation counts by the norm of the SSR in traditional algorithms). Investigating to what extent different auxiliary tasks impact the algorithm's performance, and whether simpler tasks such as predicting feature activations or parts of the input~\cite{Jaderberg17} are effective is also worth studying. Finally, it might be interesting to further investigate the connection between representation learning and exploration, since it is also known that better representations can lead to faster exploration~\cite{Jiang17}.

\section*{Acknowledgements}

The authors would like to thank Jesse Farebrother for the initial implementation of DQN used in this paper, Georg Ostrovski for the discussions and for providing us the exact results we report for \textsc{DQN$^{\scriptsize \mbox{MMC}}_{\mbox{\scriptsize{CTS}}}$} and \textsc{DQN$^{\scriptsize \mbox{MMC}}_{\mbox{\scriptsize{PixelCNN}}}$}, and Yuri Burda for providing us the data we used to compute the performance we report for RND in Atari 2600 games. We would also like to thank Carles Gelada, George Tucker and Or Sheffet for useful discussions, as well as the anonymous reviewers for their feedback. This work was supported by grants from Alberta Innovates Technology Futures and the Alberta Machine Intelligence Institute (Amii). Computing resources were provided by Compute Canada through CalculQu\'ebec. Marlos C. Machado performed this work while at the University of Alberta.

\bibliographystyle{aaai}
\bibliography{refs}

\end{document}